# Demonstration of an Aerial and Submersible Vehicle Capable of Flight and Underwater Navigation with Seamless Air-Water Transition

Marco Moreno Maia
Mechanical and Aerospace Dept.
Rutgers University
New Brunswick, NJ

Parth Soni
Mechanical and Aerospace Dept.
Rutgers University
New Brunswick, NJ

F. Javier Diez-Garias[1]
Mechanical and Aerospace Dept.
Rutgers University
New Brunswick, NJ

*Abstract*—Bio-inspired vehicles are currently leading the way in the quest to produce a vehicle capable of flight and underwater navigation. However, a fully functional vehicle has not yet been realized. We present the first fully functional vehicle platform operating in air and underwater with seamless transition between both mediums. These unique capabilities combined with the hovering, high maneuverability and reliability of multirotor vehicles, results in a disruptive technology for both civil and military application including air/water search and rescue, inspection, repairs and survey missions among others. The invention was built on a bio-inspired locomotion force analysis that combines flight and swimming. Three main advances in the present work has allowed this invention. The first is the discovery of a seamless transition method between air and underwater. The second is the design of a multi-medium propulsion system capable of efficient operation in air and underwater. The third combines the requirements for lift and thrust for flight (for a given weight) and the requirements for thrust and neutral buoyancy (in water) for swimming. The result is a careful balance between lift, thrust, weight, and neutral buoyancy implemented in the vehicle design. A fully operational prototype demonstrated the flight, and underwater navigation capabilities as well as the rapid air/water and water/air transition.

*Keywords*—bio-inspired vehicle, flight, swimming, unmanned aerial and submersible vehicle.

## I. Introduction

The development of a vehicle capable of both flight and underwater navigation has been of great interest for decades. Attempts of building such a vehicle go as far back as World War II when the Soviet Union worked on a flying submarine project [1]. More recently DARPA (2008) [2] issued a call for proposals for the development of a submersible aircraft. The interest has further grown as unmanned vehicles have seen a dramatic increase in their use for civil and military applications. This has incentivized the interest in a platform capable of air and underwater operation.

Civil applications include search and rescue, inspection, repairs and survey missions among others. The applications for performing inspection of structures such as ships hulls, underwater pier structures, oil platforms, and bridge pylons are innumerable. Additionally, beaches, rivers and lakes could be quickly surveyed and mapped for underwater erosion, oil spills, and pollution dispersion. Even life guard operation could be enhanced by having a vehicle that can operate in air and underwater when attempting rescues and reaching remote areas, as they could be used for rapid air/water deployment or as a buoyant or a towing platform. This vehicle could also be used in extreme weather (air and water) conditions where other vehicles cannot operate. Among the many military applications are launch and recovery from a submarine, rapid response to investigate a threat or a region of interest, delivering payloads to divers, rapid deployment to eliminate mines, smart/self-deploying buoy sensors, ships/ports inspections, or even stealth air/underwater missions.

There have been many bio-inspired approaches to building such a vehicle [3]. These include concepts of vehicles that could land in water and then submerge like a duck [4] or a guillemot [5] or those that would directly plunge-dive into the water like a cormorant [6] or a gannet [7]. Other approaches include vehicles that launch from underwater like a rocket [8,9], and those that can rise from underwater and glide through the air like a flying fish [10]. Bio-inspired vehicles are currently leading the way in the quest to realize such a vehicle. However, a fully functional vehicle has not yet been designed and tested. It is the complexity of operating and transitioning in both mediums (air and water) that has slowed down progress, with prototypes completing some of the tasks (maneuvers) such as landing or taking off from water, but not all that is required for a fully functional system.

Significant discoveries have been made through biological studies that focus on the mechanics of locomotion of birds and fish (flight and swim) [11,12,13,14,15,16,17]. On one hand, fish swimming locomotion has been widely studied and its propulsion classified according to which body-part generates the motion such as body-caudal fin propulsion and median-paired fin propulsion. On the other, bird flight locomotion has been classified according to the type of flight, from gliding to flapping.

The combined characteristics of swimming and of flight are only found on two groups: one is the family of flying fish [18], and the other is "aquatic birds" [19]. Aquatic birds are here loosely defined as birds that have some swimming/diving capabilities. As previously described, many vehicle concepts have tried to combine flight and swim mechanisms by imitating a specific subject in one of these two groups. Here, we propose a variation of this approach that is not based on a single animal group but rather takes the best from each group. The result will show the first tested aerial and submersible vehicle capable of flight and underwater navigation with seamless air-water transition.

---

[1] Corresponding author, diez@jove.rutgers.edu

## II. MECHANICS OF LOCOMOTION: BIO-INSPIRED FORCE ANALYSIS

Birds and fish have many features that work together enabling them to fly and swim, respectively. If we simplify the problem to the main forces acting on birds (lift, weight, thrust and drag) and on fish (buoyancy, weight, thrust, drag) we observe that the differentiators are lift and buoyancy. The discussion about forces can be further simplified by noticing that fish and also underwater mammals have a large range of weight/size ratio which indicates that weight is not a major constrain in their mechanics of locomotion. On the other hand, a bird's weight is a critical constraint in their mechanics of locomotion. We will not consider drag in our simplified analysis, as both birds and fish have mastered drag reduction through clever aerodynamics and streamlined bodies.

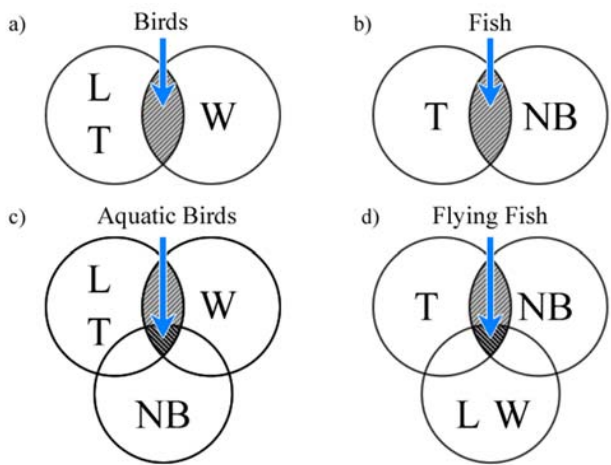

Fig. 1 - Notional sketch showing simplified forces needed for a) bird flight, b) fish swimming, c) aquatic birds flight and swimming, d) flying fish swimming and flight.

Keeping these simplifications in mind, the required forces for flying and swimming are notionally evaluated in Fig. 1. Commonly, flight locomotion requires sufficient lift and thrust for a given weight and aerodynamic shape (shaded area in Fig. 1a), while swimming locomotion requires near neutral buoyancy and sufficient thrust for a body size and aerodynamic shape (Shaded area in Fig. 1b). More revealing is how aquatic birds approach swimming and how flying fish approach flight. The former have already mastered flight. Through evolution they were able to modify their bodies to add the capability for thrust and underwater buoyancy control (shaded area in Fig. 1c). Similarly, flying fish have already mastered swimming and, through evolution, were able to modify their bodies (i.e.: enlarged pectoral fins) to add the capability for gliding in air through lift, thrust and weight control (shaded area in Fig. 1d). The key aspect is that each group has mastered one locomotion mechanism first, and added some capabilities for a secondary locomotion later. This highlights that one mechanism is more developed than the other. Birds can fly for long periods of time but can only spend a short amount of time underwater, and flying fish can swim continuously but cannot remain airborne for extended periods of time. In short, the study of animal evolution has shown us many possible mechanisms for flying and swimming, but very few that combine both. Additionally, in instances where both are combined, one mechanism clearly dominates over the other.

The question is as follows: can we combine both mechanisms (flight and swimming) into a single vehicle and be able to perform equally well in both mediums during extended periods? Notionally, this can be answered by evaluating if the shaded areas in Fig. 1a and 1b intersect. In other words, if the envelope of conditions that can produce lift and thrust for flight (for a given weight) and the envelope of conditions requiring near neutral buoyancy (in water) and thrust for swimming intersect.

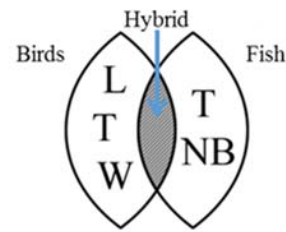

Fig. 2 - Notional sketch showing simplified forces for a proposed hybrid vehicle that would be able to take advantage of both birds flying and fish swimming capabilities.

While a comprehensive study showing this is beyond the present work, we will demonstrate that indeed when certain conditions are met, combined flight and swimming are possible, as denoted by the intersection in Fig. 2. Three main advances in the present work will show this.

## III. THE DISCOVERY OF A SEAMLESS TRANSITION METHOD BETWEEN AIR AND UNDERWATER

Among all the requirements for a vehicle to perform air and underwater operations, the transition between both mediums is perhaps the most challenging. For birds, this is done by diving directly into water or landing in water and then diving. For fixed wing vehicles, landing and taking off from water has been successfully demonstrated (seaplane). But it is the process for submersion after landing or the process for taking off after returning to the surface that adds complexity and duration to the operation. While such a vehicle has not yet been fully realized, and it wouldn't be considered a seamless transition if it did, the expectation is that such vehicle is possible and it will be fully functional in the future.

An alternative to a fixed wing vehicle is that of a single or multi-rotor vehicle. One of the advantages of multi-rotor vehicles is their VTOL (vertical take-off and landing) capability. The innovation demonstrated here redefines multirotor VTOL operations to achieve seamless transition between air and water. It uses dual propellers/motors in each vehicle-arm with a column gap between the top and bottom motors that facilitates such transition as described by the sketch in Fig. 3. Each propeller is individually addressable during the seamless transition.

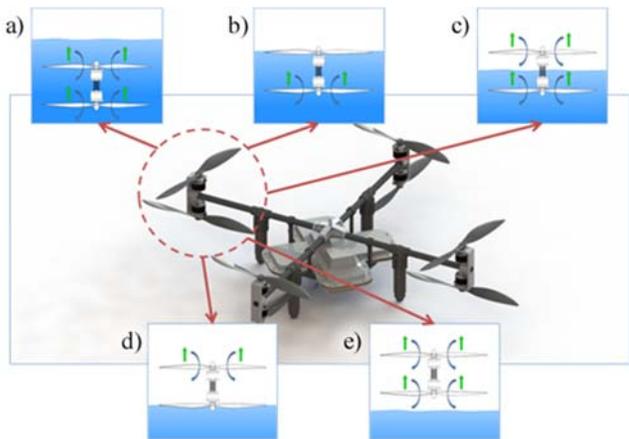

Fig. 3 - Sketch describing the mechanism used for seamless transition between water and air.

The dual-propeller system is fully reversible allowing air to underwater or underwater to air transition. The water to air transition is shown in Fig. 3. As the vehicle approaches the surface (water/air interface) both propellers (top/bottom) are generating lift. While going through the interface, the top propellers will momentarily slow down to ensure a smooth transition, and as soon as they are clear of the water, they can accelerate to entrain air and generate lift again. Similarly, as the bottom propellers reach the surface, they will momentarily slow down to ensure smooth transition, and when they clear the water interface, they can start to generate lift again. The advantage of slowing/stopping the propellers right at the air/water interface is to prevent a spike in drag from this complex interaction, where no positive forces can be generated. The sequence of events in Fig. 3a-d is performed in under two seconds, thus realizing the seamless transition. The photograph sequence in Fig. 4. shows the actual vehicle built (Naviator1) during air to water and water to air seamless transition.

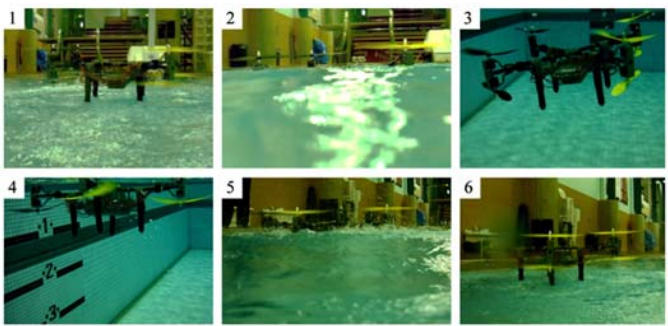

Fig. 4 - Untethered air/underwater multirotor prototype, Naviator1, performing the air to water transition (1-3) and the water to air transition (4-6).

## IV. MULTI-MEDIUM PROPULSION SYSTEM CAPABLE OF EFFICIENT OPERATION IN AIR AND UNDERWATER

Following the discussion in Figs. 1,2, the majority of a bird's propulsion system works by a flapping mechanism which generates both lift and thrust. Vehicles capable of flight need to generate these two forces. While the flapping mechanism has been difficult to reproduce beyond the smaller scales [20], two other methods are widely available. One, produced by fix wing airplanes, uses the wings to generate lift and their propellers to produce thrust. The other, produced by VTOL vehicles (single rotor and multi-rotors), uses the propellers to generate both lift and thrust and the vehicle pitching angle determines how much of the generated force is used for lift and for thrust. This capability from propellers in multi-rotor vehicles will be fully exploited here.

Fish and aquatic mammal propulsion systems can be quite diverse, but in general they all need to generate thrust. In a few species, where their body system cannot be neutrally buoyant, secondary lift forces are required. Underwater vehicles have a similar requirement to be nearly neutrally buoyant, and to generate thrust. This force is commonly generated with a propeller.

The brief discussion suggests that a propeller can be used as a multi-medium propulsion system. Two questions arise to show its feasibility:

1) Can the same propeller be used in a vertical configuration in air to generate lift and thrust, and in a horizontal configuration underwater to generate thrust?

2) If that is the case, would the performance suffer in one or the other medium?

Initially, one might not expect that the same propeller system can be able to operate in air in vertical mode and in water in horizontal mode. But as it turns out, not only it is possible but it is the preferred state. Multi-rotor vehicles in air operate in "vertical" mode where part of the force generated by the multi-rotor system is used to carry its weight. When the vehicle hovers the propellers are vertically aligned, and when the vehicle travels horizontally, the vehicle has a small pitching angle (<15-20°). A multirotor traveling at higher pitching angles can be considered to be in acrobatic mode and it is more unstable.

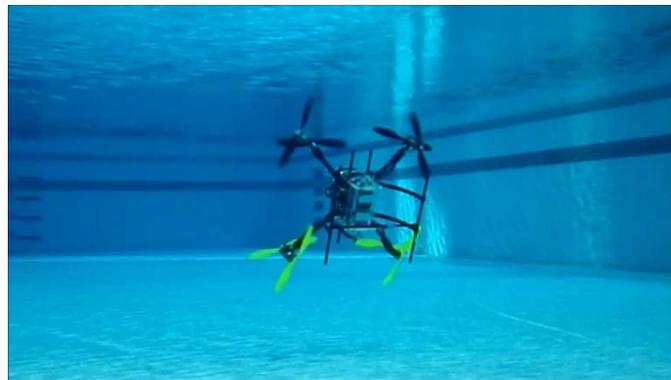

Fig. 5 - Tethered air/underwater multirotor prototype, Naviator2, during horizontal underwater cruise mode.

Multi-rotor vehicles in water have not been considered to date. Thus, their operation is not known and needs to be evaluated. As previously discussed, a horizontal-aligned propeller is desired in water, as no lift is needed in this medium for a neutrally buoyant vehicle. We have built such a vehicle to test if a multi-rotor vehicle can operate in water in "horizontal" mode. A typical image of the vehicle operating in this mode is shown in Fig. 5. The tests show this as the preferred state for multiple reasons. First, having the four dual-propellers all aligned nearly horizontally means that the majority of the force being generated is thrust. The vehicle control method is also similar to a multirotor in air, where pitch, roll and yaw are obtained by varying the power to individual arms in the vehicle. However, these are reversed due to the orientation of the vehicle. Also, for our neutrally buoyant vehicle, when travelling horizontally there is no dominant de-stabilizing force. This is critical, as such force would render horizontal mode not possible (or very unstable). In air, unfortunately, the weight of the vehicle is a de-stabilizing force that the propellers and orientation of the vehicle need to be constantly balancing to maintain control flight.

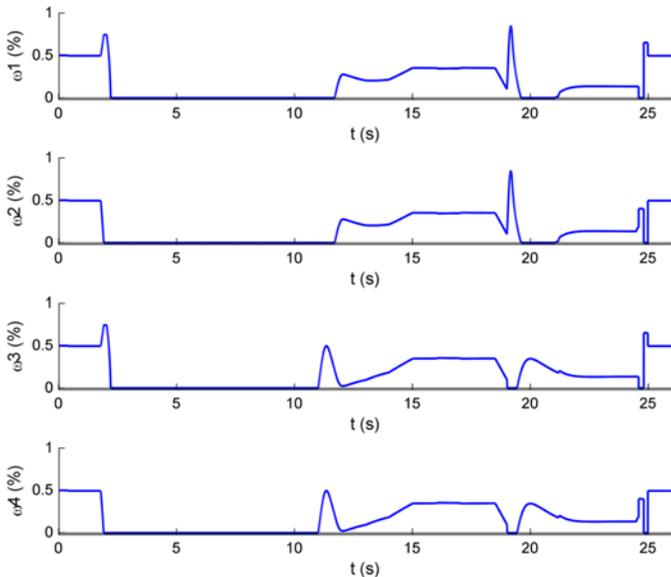

Fig. 6 – Numerical simulation of an aerial and submersible multirotor vehicle dynamics showing the angular velocity, ω1- ω4 input values to the equations of motion for rotors 1-4 as described in Appendix A.

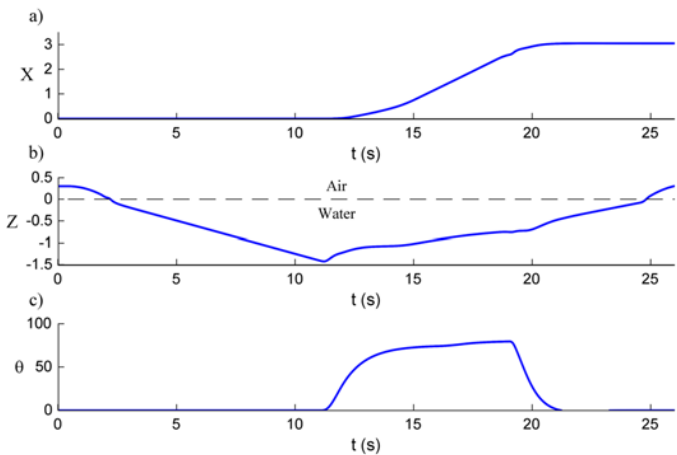

Fig. 7 – Numerical simulation of an aerial and submersible multirotor vehicle dynamics showing the numerical solutions a) X (lateral position, horizontal displacement); b) Z (vertical position, depth); c) θ (vehicle pitch angle)

The demonstrated ability of a multirotor vehicle to operate underwater opens up a new field for this widely used air platform. The first of its kind simulation of a multirotor in underwater horizontal mode is shown in Figs. 6 and 7. The full theoretical control model developed for this type of operation is described in detail in Appendix A.

The numerical solutions to the simulation in Fig. 6 and Fig. 7 were obtained using MATLAB®'s adaptive Runge-Kutta methods and demonstrate the vehicle progressing through five stages of operation. In the first stage, the vehicle crosses the air/water interface by independently controlling the rotors such that when the rotor is in the vicinity of the air/water interface, its angular velocity (or individual throttle) goes to zero. Furthermore, in order to maintain a steady descent rate, the rotors that are not near the interface increase their throttle to compensate for the thrust lost from the off or slowed rotors. Once the vehicle achieves a desired depth of 1.35m depth (typical depth used during pool testing), the second stage rotates the vehicle into horizontal mode. This utilizes a PD controller to fix the pitch and translate through the water at 65° to 75°. In the third stage, the vehicle increases its throttle to 35% to travel horizontally. In the fourth stage the vehicle rotates into vertical mode to prepare for exiting the water using the same PD controller as before. In the fifth stage, the vehicle transverses through the water/air interface while applying the same independent rotor control as in in the first stage.

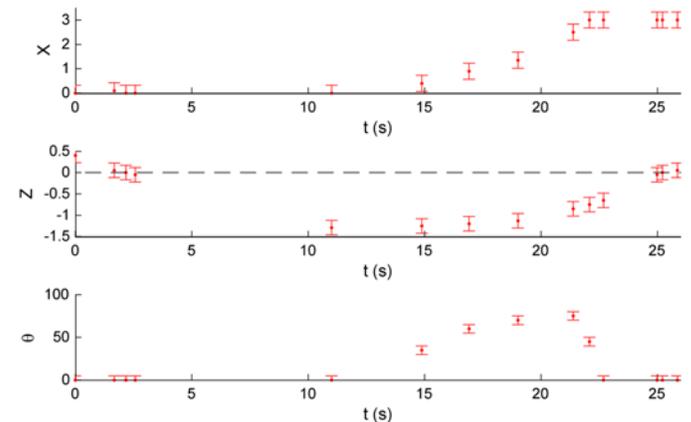

Fig. 8 – Experimental results from operation of a built prototype, "Naviator 2", during aerial and submersible operation.

The output variables (horizontal position, depth, and pith angle) from the simulation are shown in Fig. 7. They confirm that utilizing the input scheme in Fig. 6, the vehicle can transition and navigate within air and water. To validate these results, an experiment is performed with the prototype vehicle Naviator1. A mission with a very similar objective and timeframe to the simulation discussed was executed in a control pool environment. Videos of the experiment taken from different positions (outside the pool and underwater) allow extracting position, velocity and pitching angle. These results are plotted in Fig. 8 showing an excellent agreement with the numerical simulations. A video mounted with different camera shots is shown in the attached link to YouTube (https://youtu.be/XXchAsxyxjM). The small differences between the simulations and experiments are due to the fact that the vehicle was tethered for telemetry communication and the independent rotor control was not implemented. A snapshot of the test is shown in Fig. 4, where the vehicle is traveling horizontally at approximately 65-75° pitch angle.

Having shown that it is desirable to have a vertical propeller configuration in air to generate lift and thrust, and a horizontal propeller configuration underwater to generate thrust, the performance of the propellers themselves in both mediums needs to be evaluated. When considering the significant differences between air and water, the expectations are that propellers design for one medium will not operate well in the other. In general this statement is true for marine propellers. But as we will briefly show, the introduction of ultra-light weight (2-20lbs), buoyantly neutral underwater multi-rotor vehicles creates a set of requirements where the use of optimized air propellers is beneficial.

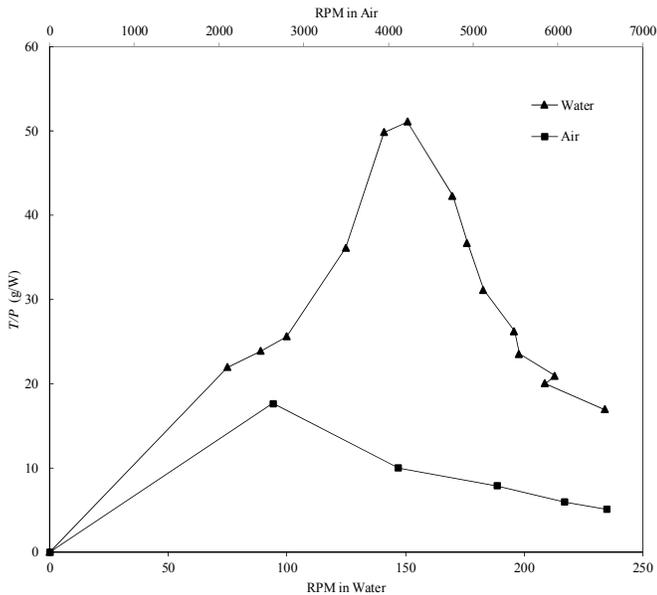

Fig. 9 – Propeller static thrust performance curves for air and water

In the design of marine propellers, a main priority is to avoid cavitation. For all the conditions tested in the present work when using optimized air propellers for underwater propulsion no cavitation was present. Briefly, the range of operating conditions for the current multirotor vehicle are: underwater cruising speeds 0.3-1m/s, with a small drag signature, requirements of 0.5-10 Newtons of thrust per propeller for maneuvering and cruising, low Reynolds number of under 100000 based on the propeller chord at 75% propeller radius, and underwater propeller RPM less than 250.

Another advantage for this type of vehicle, is that there are eight propellers generating thrust. Thus, the thrust requirements per propeller are 1/8 of what would be needed for a single propeller. This means lower RPMs, torque requirements per propeller and rotor tip velocities, and the avoidance of cavitation. The performance characteristics of a single propeller operating in air and underwater was evaluated through the use of a static load test apparatus built. It allowed measuring the static thrust at different RPM both in air and water. The results are shown in Fig. 9 for one of the propellers used.

In summary, the low thrust requirements per propeller (more typical of unmanned air vehicles than of underwater vehicles) combined with the distributed thrust requirements among eight propellers, low propeller speeds, and high underwater static thrust measurements and no cavitation, shows that air propellers are a clever choice for underwater operation. Additional advantages include the fact that they can all be used for thrust in water or in air and that each dual-propeller arm cancels its own torque which enhances the stability of the system. An alternative is to use four air propellers and four underwater propellers in each vehicle. This, while theoretically possible, would result in additional penalty in drag (four of the propellers would not be working in each medium) and the requirement that each propeller produce twice as much thrust with the corresponding increase in motor size and power consumption. Also, the torque generated in each vehicles arm would need to be cancelled by opposing arm which could affect the horizontal maneuvering underwater.

## V. QUASI-NEUTRALLY BUOYANT REQUIREMENTS

The vehicle was designed with the strict requirement of being quasi-neutrally buoyant. This generates a number of indirect constraints in weight and dimensions of the vehicle components. Using the bird analogy, evolution has shown us that birds have trimmed their weight significantly, with the largest (heaviest) known birds alive being under 16 kg (Eurasian great bustard and the African kori bustard). The reason is clear, the heavier they are the greater the amount of lift that they need to generate, which is limited by their physical capabilities. While building a vehicle we are not constraint by those physical limitations, the rule of thumb for a VTOL vehicle is to minimize weight as much as possible to reduce the size of the propulsion system. Limiting the amount of lift that needs to be generated by reducing the weight of the vehicle, increases the efficiency of the vehicle. That constraint is used for buoyancy requirements. This is controlled by properly sizing a buoyancy tank where most of the vehicle components are hosted. In general, the smaller the weight of the vehicle, the smaller the buoyancy tank needed. Ideally, the volume of the buoyancy tank is as small as the volume of the components it contains. Anything bigger adds weight which adds lift requirements in air. For large buoyancy requirements, the vehicle becomes too heavy and impractical for flight. Surprisingly when all the components are considered and miniaturized when possible, the final vehicle weight (<2kg) and size (<0.9m span) nicely scales with those of many flying birds.

## VI. CONCLUSION

A brief bio-inspired locomotion analysis suggest that for a vehicle to show similar performance in both air and water, the envelope of conditions that can produce lift and thrust for flight (for a given weight) and the envelope of conditions requiring near neutral buoyancy (in water) and thrust for swimming need to overlap. Such vehicle was realized during this work through three main advances.

Among all the requirements for a vehicle to be able to perform both air and underwater missions, the transition between both mediums is perhaps the most challenging. An innovative dual propeller system demonstrated this seamless transition. Briefly, as the vehicle approaches the water/air interface both propellers are generating lift. While going through the interface, the top propellers will momentarily slow down to ensure a smooth transition, and as soon as they are clear of the water, they can accelerate to entrain air and generate lift again. Similarly, as the bottom propellers reach the surface, they will momentarily slow down to ensure smooth transition, and when they clear the water interface, they can start to generate lift again.

Recognizing that flight requires both lift and thrust generation, and swimming requires thrust generation, a multi-medium propulsion system was proposed. In air, it uses a vertical propulsion mode that generates both lift and thrust. In water, it uses a horizontal propulsion mode that generates thrust. We show the advantages of using air propellers for both the air and underwater propulsion mode. The propeller static thrust tests show similar performance of these propellers in both mediums.

The demonstrated capabilities of a multirotor vehicle to operate underwater opens up a new field for this widely used air platform. A full theoretical control model capable of handling both mediums is developed for this platform showing excellent agreement with experimental tests.

Designing and building a vehicle that is neutrally buoyant is a requirement that affects all the systems in the vehicle. It generates constraints in terms of weight and size. In exchange, a neutrally buoyant vehicle, such as the prototype built, is able to perform extended missions underwater while minimizing the underwater propulsion requirements.

**Acknowledgements** This work is supported by the Office of Naval Research under Award No.: N00014-15-1-2235

REFERENCES

[1] Paul Marks, "From sea to sky: Submarines that fly," *New Scientist*, July 2010.
[2] "Broad Agency Announcement: Submersible Aircraft," DARPA, DARPA-BAA-09-06, 2008.
[3] R. Siddall and M. Kovač, "Launching the AquaMAV: bioinspired design for aerial–aquatic robotic platforms," *Bioinsp. Biomim.*, vol. 9, no. 3, 2014.
[4] T. Aigeldinger and F. Fish, "Hydroplaning by ducklings: overcoming limitations to swimming at the water surface," *Experimental Biology, Journal of*, vol. 198, pp. 1567-1574, July 1995.
[5] R. J. Lock, R. Vaidyanathan, S. C. Burgess, and J. Loveless, "Development of a biologically inspired multi-modal wing model for aerial-aquatic robotic vehicles through empirical and numerical modelling of the common guillemot, Uria aalge," *Bioinspir. Biomim.*, vol. 5, no. 4, 2010.
[6] Terrence A. Weisshaar, "Morphing aircraft systems: historical perspectives and future challenges," *Journal of Aircraft*, vol. 50, no. 2, pp. 337-353, 2013.
[7] A. Fabian, Y. Feng, E. Swartz, D. Thurmer, and R. Wang, "Hybrid Aerial Underwater Vehicle (MIT Lincoln Lab)," *2012 SCOPE Projects*, Paper 8 (2012), http://digitalcommons.olin.edu/scope_2012/8/.
[8] Kollmorgen, "Sea Sentry Organic Submarine Launched UAV," Kollmorgen Corporation, Radford, 2009.
[9] D. Majumdar, "U.S. Navy Launches UAV from a Submarine," *U.S. Naval Institute News*, 2013.
[10] A. Lussier Desbiens, M. T. Pope, D. L. Christensen, E. W. Hawkes, and M. R. Cutkosky, "Design principles for efficient, repeated jumpgliding," *Bioinspir. Biomim.*, vol. 9, no. 2, pp. 1-12, 2014.
[11] C. M. Breder, "The locomotion of fishes.," *Zoologica*, vol. 50, pp. 159–297, 1926.
[12] P. W. Webb, "Swimming," in *The Physiology of Fishes*, 2nd ed., D. H. Evans, Ed. New York: CRC Press, 1998, pp. 3-24.
[13] Michael Sfakiotakis, David M. Lane, J. Bruce, and C. Davies, "Review of fish swimming modes for aquatic locomotion," *IEEE Journal of Oceanic Engineering*, vol. 24, no. 2, pp. 237-252, 1999.
[14] B. W. Tobalske, "Comparative power curves in bird flight.," *Nature*, vol. 421, no. 6921, pp. 363-366, 2003.
[15] K. P. Dial, "Mechanical power output of bird flight.," *Nature*, vol. 390, no. 6655, pp. 67-70, 1997.
[16] B. W. Tobalske, "Biomechanics of bird flight.," *Experimental Biology, Journal of*, vol. 210, no. 18, pp. 3135-3146, 2007.
[17] Colin J. Pennycuick, *Bird Flight Performance*.: Oxford University Press, 1989.
[18] J. Davenport, "How and why do flying fish fly?," *Rev. Fish Biol. Fish.*, vol. 4, no. 2, pp. 184-214, 1994.
[19] J. R. Lovvorn and D. R. Jones, "Body mass volume, and buoyancy of some aquatic birds, and their relation to locomotor strategies," *Canadian Journal of Zoology*, vol. 69, pp. 2888-2892, 1991.
[20] T. Nakata et al., "Aerodynamics of a bio-inspired flexible flapping-wing micro air vehicle," *Bioinsp. Biomim.*, vol. 6, no. 045002, 2001.
[21] Haim Baruh, "Three-Dimensional Kinematics of Rigid Bodies," in *Applied Dynamics*. New York, United States: CRC Press, 2015, ch. 9.
[22] S. Bouabdallah, P. Murrieri, and R. Siegwart, "Design and control of an indoor micro quadrotor," *Proc. IEEE Int. Conf. Robotics and Automation (ICRA)*, vol. 5, Apr. 26–May 1, 2004.
[23] T. Hamel, R. Mahony, R. Lozano, and J. Ostrowski, "Dynamic modelling and configuration stabilization for an X4-flyer," *Proc. Int. Federation of Automatic Control Symp. (IFAC)*, 2002.
[24] J. Seddon, *Basic Helicopter Aerodynamics*. Oxford: BSP Professional Books, 1990.
[25] Colin P. Coleman, "A Survey of Theoretical and Experimental Coaxial Rotor Aerodynamic Research," *NASA Technical Paper*, 1997.
[26] Péter Fankhauser, Samir Bouabdallah, Stefan Leutenegger, and Roland Siegwart, "Modeling and Decoupling Control of the CoaX Micro Helicopter," *IEEE/RSJ International Conference on Intelligent Robots and Systems*, pp. 2223-2228, September 2011.
[27] R. Mahony, V. Kumar, and P. Corke, "Multirotor Aerial Vehicles: Modeling, Estimation, and Control of Quadrotor," *Robotics & Automation Magazine, IEEE*, vol. 19, no. 3, pp. 20 - 32, August 2012.
[28] J. S. Rao, "Gyroscopic Effects," in *Rotor Dynamics*. New Delhi, India: New Age International (P) Ltd., 1996, ch. 6, p. 135.

APPENDIX A

The vehicle platform is an eight rotor vehicle based on the octa-quadcopter configuration, which is a multirotor with four arms where each arm consists of two counter-rotating rotors along a single axis. A set of one or more pressure vessels along the vertical central axis of the vehicle houses the water-sensitive components and controls the angular velocities of the brushless motors.

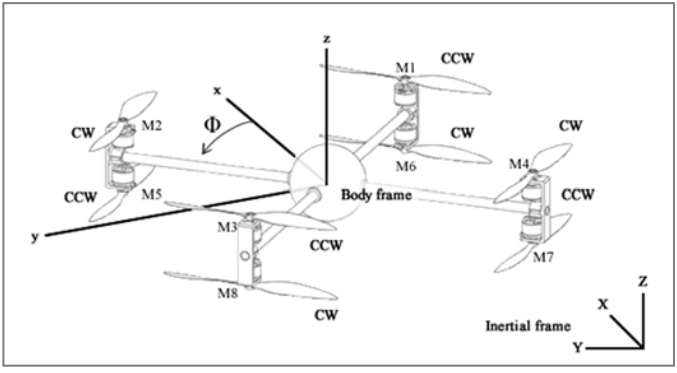

Fig. 10 – Sketch of an aerial and submersible multirotor vehicle showing the body frame and inertial frame coordinates and location of the rotors.

As seen on Fig. 10, each arm is offset by an angle $\Phi$ about the body frame z-axis and each rotor is assigned a reference number and aerial spin direction. For convenience and in accordance with Fig. 10, let the uppercase $XYZ$ axis represent the inertial frame and the lowercase $xyz$ axis represent the body frame with unit vectors $\{\hat{I},\hat{J},\hat{K}\}$ and $\{\hat{i},\hat{j},\hat{k}\}$, respectively. Using Euler angles, we can use the rotation matrix $[R]$ to relate the body frame to the inertial frame using a 3-2-1 (or Z-X-Y) Euler angle sequence rotation [21] using the Euler angles $\psi$ (yaw), $\theta$ (pitch) and $\phi$ (roll), respectively:

$$[R] = [R_\phi][R_\theta][R_\psi] \qquad (1)$$

$$[R] = \begin{bmatrix} c\psi c\theta & c\theta s\psi & -s\theta \\ c\psi s\theta s\phi - c\phi s\psi & s\psi s\theta s\phi + c\phi c\psi & c\theta s\phi \\ c\psi c\phi s\theta + s\psi s\phi & c\phi s\psi s\theta - s\phi c\psi & c\theta c\phi \end{bmatrix} \qquad (2)$$

Where the rotation from $xyz$ to $XYZ$ is given by

$$[R]^T = \begin{bmatrix} c\psi c\theta & c\psi s\theta s\phi - c\phi s\psi & c\psi c\phi s\theta + s\psi s\phi \\ c\theta s\psi & s\psi s\theta s\phi + c\phi c\psi & c\phi s\psi s\theta - s\phi c\psi \\ -s\theta & c\theta s\phi & c\theta c\phi \end{bmatrix} \qquad (3)$$

The dynamics model used will be the Newton-Euler formalism [22]:

$$\begin{bmatrix} mI_{3x3} & 0 \\ 0 & I \end{bmatrix} \begin{pmatrix} \dot{v} \\ \dot{\omega} \end{pmatrix} + \begin{bmatrix} \omega \times mv \\ \omega \times I\omega \end{bmatrix} = \begin{pmatrix} F \\ \tau \end{pmatrix} \qquad (4)$$

In (4) $I$ is the inertia matrix ($\in \mathbb{R}^{3x3}$), $v$ is the translational velocity vector, $m$ is the total mass of the vehicle, $\omega$ is the angular velocity vector, $F$ is the forces vector and $\tau$ is the moments vector. We can separate (4) into the form [23]:

$$\dot{r} = v \qquad (5)$$

$$m\dot{v} = [R]^T F = -mg\hat{K} + [R]^T F_{nc} \qquad (6)$$

$$I\dot{\omega} = -\omega \times I\omega + \tau \qquad (7)$$

Where
- $r$ — position of the rigid body's centroid $\in$ inertial frame
- $\tau_i$ — motor torque input
- $F$ — forces in the body frame
- $\tau$ — moments in the body frame
- $F_{nc}$ — non-conservative forces in the body frame
- $I$ — inertia matrix for the rigid body ($\in \mathbb{R}^{3x3}$)
- $I_r$ — inertia matrix for the rotor ($\in \mathbb{R}^{3x3}$)
- $\sigma_i$ — rotor spin direction $\in \{-1:CCW, +1:CW\}$
- $\Delta$ — secondary forces that are induced when not hovering

Using momentum theory [24], the relationship between thrust and rotor angular velocity during hover can be described as,

$$C_T = T / \rho A (\varpi b)^2 \qquad (8)$$

Where $T$ is the thrust, $\rho$ is the density, $A$ is rotor disk area, $\varpi$ is the rotor angular velocity and b is the rotor blade length or rotor radius. Since the platform in this scope comprises of equally sized rotors, (8) may be simplified to (9). Similarly as in, the reaction torque can be modeled as [23].

$$T = K_T \rho \varpi^2 \qquad (9)$$

$$Q = K_Q \rho \varpi^2 \qquad (10)$$

It is true that the top rotors affect the performance of the bottom rotors and thus, the relationship between thrust and rotor angular velocity during hover is better approximated by [25],

$$C_T = \left(T_{upper} + T_{lower}\right) / \rho A (\varpi b)^2 . \qquad (11)$$

However, since only small disparities in upper and lower rotor angular velocities are necessary to achieve yaw, this interference may be neglected [26]. The total thrust generated for hover is given by [27]:

$$T_\Sigma = \sum_{i=1}^{8} T_i = \sum_{i=1}^{8} K_T \rho_i \varpi_i^2 = K_T \sum_{i=1}^{8} \rho_i \varpi_i^2 \qquad (12)$$

If rotor flapping is neglected we can define the non-conservative forces as (13), where $T_\Sigma$ acts in the positive z-axis direction $\hat{k}$ and $\Delta$ is a placeholder for the secondary aerodynamic forces that are induced when not hovering. The non-conservative forces in the body frame can be written as,

$$F_{nc} = T_\Sigma \hat{k} + \Delta . \qquad (13)$$

The torque vector $\tau$ may then be defined as

$$\tau = \begin{pmatrix} \tau_\phi & \tau_\theta & \tau_\psi \end{pmatrix}^T = \tau_a - \tau_b , \qquad (14)$$

where $\tau_a$ is the moment due to obvious reactions and $\tau_b$ is the resistive moment due to changes in the rigid body's orientation (the gyroscopic effect) [28]:

$$\tau_a = \begin{pmatrix} K_T \sum_{i=1}^{8} \rho_i d \sin(\Phi_i) \varpi_i^2 \\ -K_T \sum_{i=1}^{8} \rho_i d \cos(\Phi_i) \varpi_i^2 \\ K_Q \sum_{i=1}^{8} \rho_i \sigma_i \varpi_i^2 \end{pmatrix} \quad (15)$$

$$\tau_b = I_r \sum_{i=1}^{8} (\omega \times \hat{k}) \varpi_i \quad (16)$$

In (15), $d$ is the arm length from the centroid to the rotor in the xy-plane and in (16), the difference in moment of inertia for the CCW and CW rotors are taken to be negligible. The density $\rho_i \in \{\rho_{upper}, \rho_{lower}\}$, whereby $\rho_{upper}$ is the density of the space above the two-medium interface and $\rho_{lower}$ is the density of the space below the two-medium interface. The platform discussed is designed to be a vehicle capable of operating in air and water and as such, the interface of interest is the air-water interface. Therefore, the density can be given as

$$\rho_i \begin{cases} \rho_{air} & \text{if sensors indicate rotor in air} \\ \rho_{water} & \text{if sensors indicate rotor in water} \end{cases} \quad (17)$$

A Heaviside function, $H$, can be used to implement this behavior, where $\rho_i$ becomes (18) where $Z$ is the airframe position relative to the air-water interface in the $\hat{K}$ direction and $h_i$ is the rotor distance relative to the airframe. The density is then given by

$$\rho_i = \rho_{water} + H(Z - h_i)(\rho_{air} - \rho_{water}) \quad (18)$$

The three translation equations of motion become,

$$m\ddot{X} = [R]^T (T_\Sigma \hat{k} + \Delta) \cdot \hat{I}, \quad (19)$$

$$m\ddot{Y} = [R]^T (T_\Sigma \hat{k} + \Delta) \cdot \hat{J}, \quad (20)$$

$$m\ddot{K} = -mg + [R]^T (T_\Sigma \hat{k} + \Delta) \cdot \hat{K}. \quad (21)$$

A generic $I$ is defined such that the platform's geometry yields

$$I = \begin{bmatrix} I_{xx} & 0 & 0 \\ 0 & I_{yy} & 0 \\ 0 & 0 & I_{zz} \end{bmatrix}. \quad (22)$$

We apply Euler's equations [21,22], which can be expressed in vector form as (23) or (24) and can be further expanded to (25).

$$I\dot{\omega} + \omega \times (I\omega) = \tau \quad (23)$$

$$\dot{\omega} = \begin{pmatrix} \ddot{\psi} \\ \ddot{\theta} \\ \ddot{\phi} \end{pmatrix} = I^{-1}(\tau - \omega \times (I\omega)) \quad (24)$$

$$\begin{pmatrix} \ddot{\phi} \\ \ddot{\theta} \\ \ddot{\psi} \end{pmatrix} = \begin{pmatrix} \tau_\phi I_{xx}^{-1} \\ \tau_\theta I_{yy}^{-1} \\ \tau_\psi I_{zz}^{-1} \end{pmatrix} + \begin{pmatrix} \dfrac{I_{yy} - I_{zz}}{I_{xx}} \dot{\theta}\dot{\psi} \\ \dfrac{I_{zz} - I_{xx}}{I_{yy}} \dot{\psi}\dot{\phi} \\ \dfrac{I_{xx} - I_{yy}}{I_{zz}} \dot{\theta}\dot{\phi} \end{pmatrix} \quad (25)$$

The three equations of motion obtained from Euler's equation are,

$$\ddot{\phi} = \frac{I_{yy} - I_{zz}}{I_{xx}} \dot{\theta}\dot{\psi} + \tau_\phi I_{xx}^{-1} \quad (26)$$

$$\ddot{\theta} = \frac{I_{zz} - I_{xx}}{I_{yy}} \dot{\psi}\dot{\phi} + \tau_\theta I_{yy}^{-1} \quad (27)$$

$$\ddot{\psi} = \frac{I_{xx} - I_{yy}}{I_{zz}} \dot{\theta}\dot{\phi} + \tau_\psi I_{zz}^{-1} \quad (28)$$

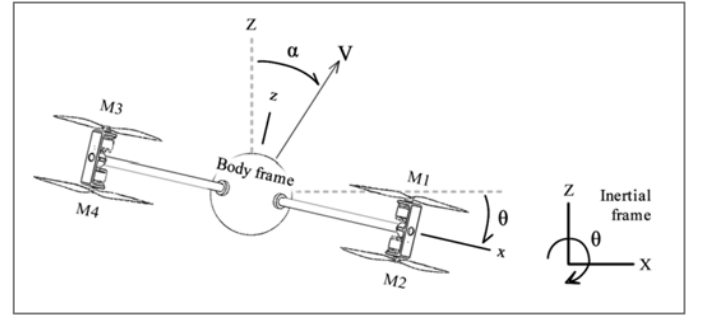

Fig. 11 – 2-Dimensional model of an aerial and submersible multirotor vehicle showing the body frame and inertial frame coordinates, the angle of attack and location of the rotors.

We are interested in transition behavior and so, we consider the two-dimensional case with the very simplified equations, which incorporate drag into $\Delta$, where $\alpha$ is the angle of attack calculated from the velocity vector and the $\hat{K}$ unit vector. The simplified model of the vehicle platform is shown in Fig. 11. The constants to be used in the simulation are listed in Table 1.

$$\ddot{X} = \frac{\sin \theta}{m} [2 K_T (\rho_1 \varpi_1^2 + \rho_2 \varpi_2^2 + \rho_3 \varpi_3^2 + \rho_4 \varpi_4^2)] + \frac{F_d \sin \alpha}{m} \quad (29)$$

$$\ddot{Z} = -g + \frac{\cos \theta}{m} [2 K_T (\rho_1 \varpi_1^2 + \rho_2 \varpi_2^2 + \rho_3 \varpi_3^2 + \rho_4 \varpi_4^2)] + \frac{F_d \cos \alpha}{m} \quad (30)$$

$$\ddot{\theta} = 2 K_T d \frac{\sqrt{2}}{2} (-\rho_1 \varpi_1^2 - \rho_2 \varpi_2^2 + \rho_3 \varpi_3^2 + \rho_4 \varpi_4^2) I_{yy}^{-1} \quad (31)$$

TABLE I. SIMULATION PROPERTIES AND PARAMETERS

| Constants | Value | Units | Constants | Value | Units |
|---|---|---|---|---|---|
| $m$ | 2.00 | $kg$ | $\rho_{air}$ | 1.225 | $kg/m^3$ |
| $I_{yy}$ | $3.46 \cdot 10^{-2}$ | $kg \cdot m^2$ | $\rho_{water}$ | 999.97 | $kg/m^3$ |
| $K_t$ | $1.34 \cdot 10^{-5}$ | $m^4/rad^2$ | $g_{air}$ | 9.81 | $m/s^2$ |
| $d$ | 0.3 | $m$ | $g_{water}$ | 0.35 | $m/s^2$ |
| $C_d$ | 0.8 | | $\omega_{max_{air}}$ | 773.1 | $rad/s$ |
| $A$ | $6.16 \cdot 10^{-2}$ | $m^2$ | $\omega_{max_{water}}$ | 23.25 | $rad/s$ |
| $K_{P2}$ | 1.00 | | $K_{P1}$ | 1.50 | |
| $K_{D2}$ | -0.84 | | $K_{D1}$ | -1.70 | |